\documentclass[runningheads]{llncs}

\usepackage{eccv}

\usepackage{eccvabbrv}

\usepackage{booktabs}
\usepackage{graphicx}
\usepackage{amsmath,amssymb} 
\usepackage{color}
\usepackage{subcaption}
\usepackage{multirow}
\usepackage{adjustbox}
\usepackage{bbm}
\usepackage{booktabs}
\usepackage{wrapfig}
\usepackage[ruled]{algorithm2e}

\usepackage{hyperref}

\begin{document}

\title{Improving 3D Semi-supervised Learning by Effectively Utilizing All Unlabelled Data} 

\titlerunning{Improving 3D SSL by Effectively Utilizing All Unlabelled Data}

\author{Sneha Paul \and
Zachary Patterson \and
Nizar Bouguila}

\authorrunning{S.~Paul et al.}

\institute{Concordia Institute for Information Systems Engineering, \\
Concordia University, Montreal, Canada \\
\email{sneha.paul@mail.concordia.ca}\\
\email{\{zachary.patterson,nizar.bouguila\}@concordia.ca}}

\maketitle

\begin{abstract}
  Semi-supervised learning (SSL) has shown its effectiveness in learning effective 3D representation from a small amount of labelled data while utilizing large unlabelled data. Traditional semi-supervised approaches rely on the fundamental concept of predicting pseudo-labels for unlabelled data and incorporating them into the learning process. However, we identify that the existing methods do not fully utilize all the unlabelled samples and consequently limit their potential performance. To address this issue, we propose AllMatch, a novel SSL-based 3D classification framework that effectively utilizes all the unlabelled samples. AllMatch comprises three modules: (1) an adaptive hard augmentation module that applies relatively hard augmentations to the high-confident unlabelled samples with lower loss values, thereby enhancing the contribution of such samples, (2) an inverse learning module that further improves the utilization of unlabelled data by learning what not to learn, and (3) a contrastive learning module that ensures learning from all the samples in both supervised and unsupervised settings. Comprehensive experiments on two popular 3D datasets demonstrate a performance improvement of up to 11.2\% with 1\% labelled data, surpassing the SOTA by a significant margin. Furthermore, AllMatch exhibits its efficiency in effectively leveraging all the unlabelled data, demonstrated by the fact that only 10\% of labelled data reaches nearly the same performance as fully-supervised learning with all labelled data. The code of our work is available at: \href{https://github.com/snehaputul/AllMatch}{github.com/snehaputul/AllMatch}.

  \keywords{Semi-supervised Learning \and Point Cloud \and 3D Vision}
\end{abstract}

\vspace{-20pt}
\section{Introduction}
\vspace{-7pt}
\begin{wrapfigure}{r!}{0.38\textwidth}
\vspace{-67pt}
\begin{center}
    \centering
    \includegraphics[width=1.0\linewidth]{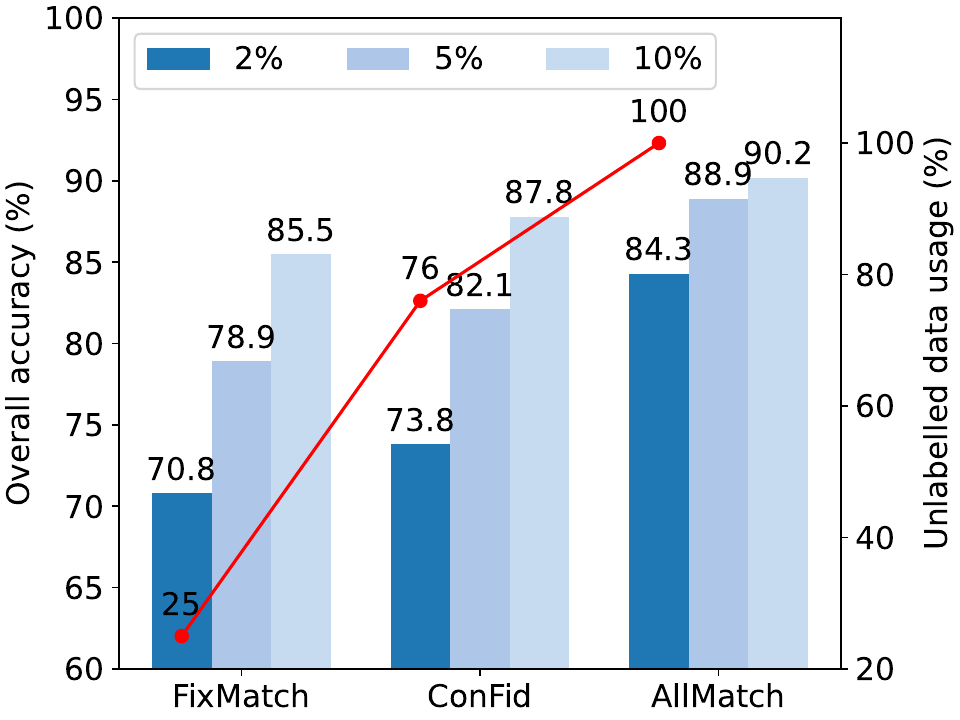}
    \caption{Performance in comparison to unlabelled data usage.}
    \label{fig:banner}
\end{center}
\vspace{-50pt}
\end{wrapfigure}

In the realm of 3D computer vision, Point Cloud applications are on the rise, finding extensive use in diverse domains. This increasing popularity of Point Cloud applications across various 3D computer vision domains has heightened the demand for annotated data, a resource-intensive and expensive task \cite{liu2022point}. To tackle this issue, 3D semi-supervised learning (SSL) emerges as a promising avenue. By capitalizing on scarce labelled data in conjunction with abundant unlabelled data, this approach has demonstrated its efficacy in different computer vision applications, including 3D object recognition. Most popular forms of SSL (e.g. \cite{fixmatch}, \cite{flexmatch}) involve predicting pseudo-labels for the unlabelled data and using \textit{only} the high-confident predictions for learning. However, due to the hand-designed nature of the high-confidence-based filtering, most of the unlabelled samples never contribute to the learning (Figure \ref{fig:banner}).

\begin{wrapfigure}{r!}{0.5\textwidth}
    \vspace{-25pt}
    \centering
     \begin{subfigure}[b]{0.23\textwidth}
         \centering
         \includegraphics[width=1.\textwidth]{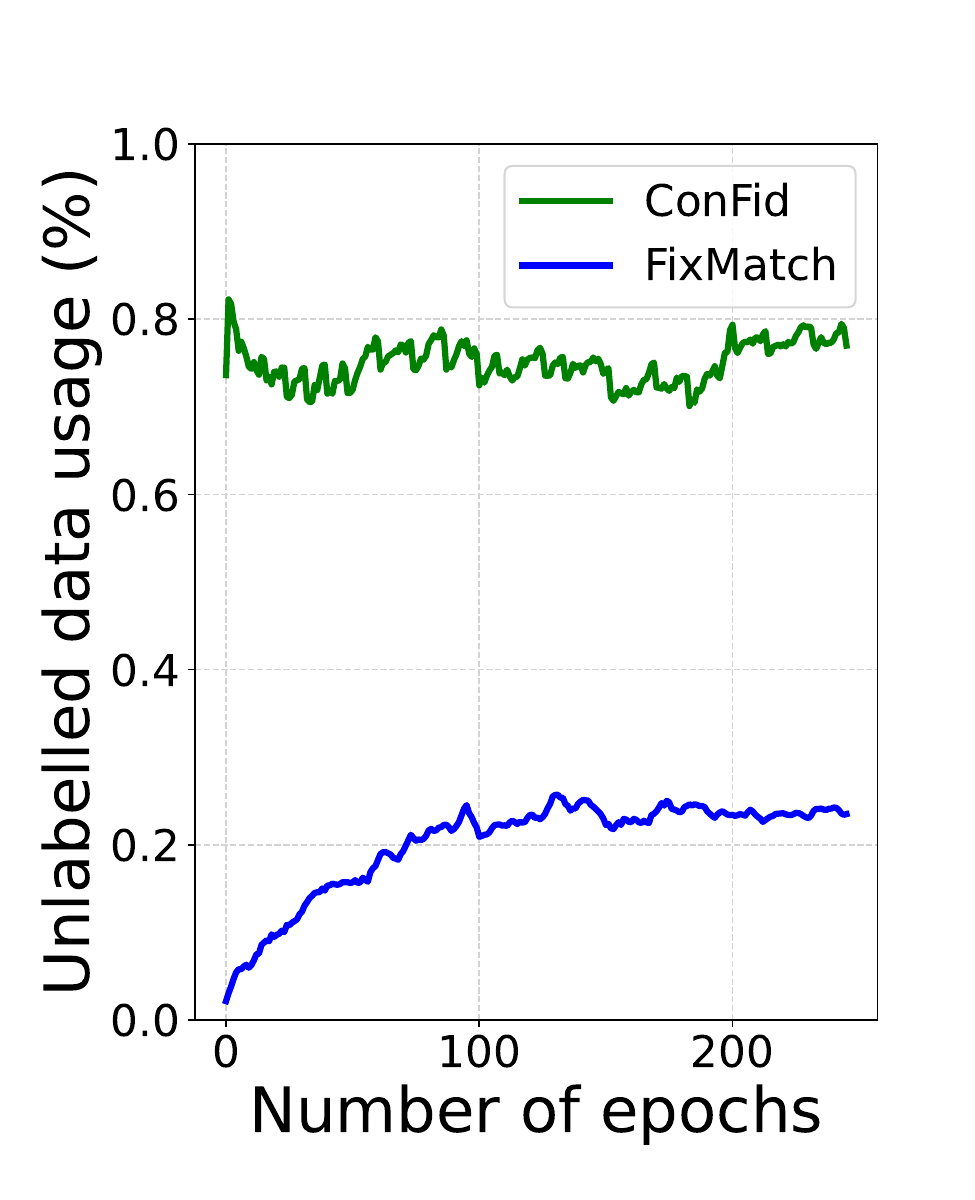}
     \end{subfigure}
      \begin{subfigure}[b]{0.23\textwidth}
         \centering
         \includegraphics[width=1\textwidth]{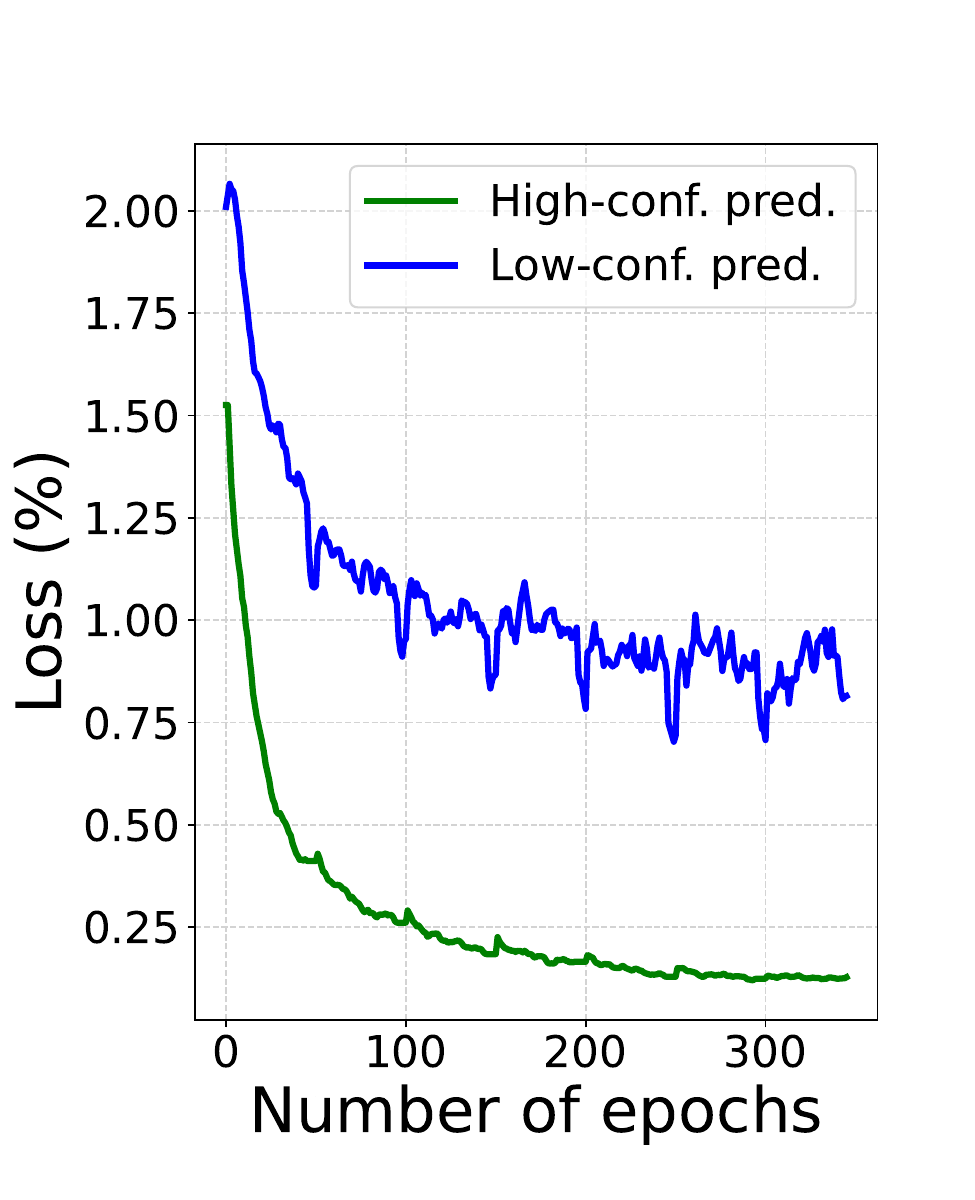}
     \end{subfigure}
    \vspace{-1em}
    \caption{(left) Pseudo label utilization over the training epochs by FixMatch and ConFid. While ConFid does better utilization than FixMatch, none properly utilizes all unlabelled data.
    (right) Loss over training for high and low-confident pseudo-labels. The high-confident samples are selected for unsupervised learning, but low loss results in marginal contribution to learning.
    }
    \label{fig:pl_usage}
\vspace{-20pt}
\end{wrapfigure}

Some of the more recent works propose different techniques to increase the number of unlabelled samples the model learns from. For example, FlexMatch \cite{flexmatch} introduced a dynamic threshold strategy, where the threshold value is adjusted according to the model's learning status using curriculum learning. SoftMatch \cite{softmatch} proposed a trade-off between quality and quantity to increase unlabelled data usage. While there has been some progress in this direction, these methods are specially developed for the image domain and have not been explored in the context of 3D SSL. Furthermore, as shown in Figure \ref{fig:pl_usage} (left) for 3D SSL, FixMatch utilizes about 25\% of the unlabelled data, while the current state-of-the-art (SOTA) method, ConFid \cite{confid}, utilizes about 76\% of the unlabelled data. Moreover, we find that the correctly predicted pseudo-labels do not contribute much towards the learning, as they have higher confidence in prediction and lower loss (Figure \ref{fig:pl_usage} (right)).

In this work, we propose a novel semi-supervised framework named AllMatch that handles the above-mentioned problems and ensures learning from \textit{all} the unlabelled data. AllMatch consists of three modules, all of which improve different aspects of increasing unlabelled data usage. While the concepts of these modules are not entirely new, they have never been explored in 3D SSL, especially with the focus on increasing the utilization of unlabelled data. The modules in AllMatch are: (a) An adaptive hard augmentation(AHA) module that increases the augmentation strength of the high-confident unlabelled samples with lower loss. This module ensures better utilization of high-confident samples that the model has already learned, thus leaving no further contribution to the model's learning process. (b) An inverse learning module that further ensures better utilization of unlabelled samples by assigning inverse pseudo labels to the low-confident predictions and learning \textit{what not to learn}. 
(c) A contrastive learning (CL) module that learns from all the unlabelled data in unsupervised learning settings. This module ensures that the samples excluded from pseudo-label-based learning also contribute to the learning process by participating in CL. Additionally, we utilize a modified supervised contrastive loss \cite{supcon} for learning from the labelled data in a similar learning setting as the unlabelled data.

We evaluate AllMatch on two popular point cloud datasets, ModelNet40 \cite{modelnet}, and ScanObjectNN \cite{scanobject}, with different amounts of labelled sets. We compare AllMatch with the existing SSL methods and find that AllMatch outperforms the existing SOTA by up to 11.2\% and 10.45\% on ScanObjectNN and ModelNet40 datasets, respectively, with a minimal amount of labelled data. AllMatch also shows superior performance with a lower amount of labelled set, indicating its effectiveness in utilizing all unlabelled samples. With our proposed method, only 10\% of the labelled data performs almost as well as in a fully supervised setting with all the labelled data. Furthermore, our method is more computationally efficient than the previous SOTA, ConFid \cite{confid}, which requires 500 epochs of training compared to 250 in ours.
We also conduct an in-depth ablation study and sensitivity analysis demonstrating the importance of all three components for AllMatch's superior performance. Here, the AHA module shows the highest impact on the model's performance, followed by inverse learning and contrastive learning modules. Our main contributions to this work can be summarized as follows:

\begin{itemize}
    \item We propose a novel semi-supervised 3D classification framework named AllMatch that effectively utilizes the whole unlabelled set to improve the performance. 
    \item We introduce an adaptive hard augmentation module that efficiently utilizes the higher-confident samples with lower loss and an inverse learning module to further boost the utilization of unlabelled samples by learning from low-confident predictions. 
    \item We introduce a contrastive learning module to ensure learning from the remainder of the unlabelled data that are not utilized by pseudo-label-based learning.
    \item We improve the SOTA for 3D semi-supervised learning by 11.2\%, and 10.45\% on the ScanobjectNN and ModelNet40 datasets, respectively. We also show that AllMatch achieves almost similar results in a fully-supervised setting with only 10\% of labelled data.
    
\end{itemize}

\section{Related Work}
\subsection{Semi-supervised Learning}
There has been significant progress in semi-supervised learning in recent years, mostly for image representation learning. Most dominant approaches in semi-supervised learning can be divided into two main categories: consistency regularization \cite{mean_teacher,uda} and entropy minimization \cite{fixmatch,remixmatch,mixmatch}. The consistency regularization learns by enforcing consistency in the prediction of the model under different perturbations. On the other hand, entropy minimization learns by reducing the entropy in the prediction of the model on the unlabelled data. One of the most prominent methods in this category is FixMatch \cite{fixmatch}. The basic concept of FixMatch is to predict the pseudo-label for the unlabelled data from a weakly augmentation of the input and then utilize this as the supervision for a strongly augmented sample if the confidence of the pseudo-label is high. One main issue with this approach is its reliance on the quality of the pseudo-labels. Also, the fixed thresholding nature results in very low utilization of the unlabelled data as the low confidence prediction does not contribute towards the learning process. 
Recently, several methods have been proposed to increase the utilization of the unlabelled data by proposing variants of the thresholding concept, including Dash \cite{dash}, FlexMatch \cite{flexmatch}, CoMatch \cite{comatch}, SimMatch \cite{simmatch}, ConMatch \cite{conmatch}.
 
Dash \cite{dash} selects unlabelled samples based on a criterion where their cross-entropy loss is smaller than an adjustable threshold value. 
FlexMatch \cite{flexmatch} is another popular work that utilizes curriculum learning to adapt the threshold value, taking into account the class-wise learning status. This dynamic threshold is then applied to select pseudo-labels with high confidence.
CoMatch \cite{comatch} introduces a co-training framework that involves the interaction between class probabilities and embeddings. The embeddings impose a smoothness constraint on class probabilities to enhance the quality of pseudo-labels. These refined pseudo-labels are used as targets for training both the classification head with cross-entropy loss and the projection head with a graph-based contrastive loss.
SimMatch \cite{simmatch} adeptly aligns similarity relationships at dual levels, utilizing a memory buffer for annotated examples to amplify the synergy between semantic and instance pseudo-labels.
ConMatch \cite{conmatch} elevates model efficacy through the integration of confidence-guided consistency regularization. This approach tackles the hurdle of learning from a restricted pool of labelled data by promoting consistent predictions on unlabelled data, all the while taking into account the confidence levels associated with those predictions.
SequenceMatch \cite{sequencematch} enhances semi-supervised learning by minimizing prediction distribution divergence between weakly and strongly augmented examples. It underscores the significance of diverse augmentations and consistency constraints in the learning process.
CHMatch \cite{chmatch} integrates instance-level prediction matching and contrastive graph-level matching through the use of a memory-bank-based adaptive threshold strategy. It employs hierarchical label-guided graph matching to enhance contrastive feature learning, providing a more robust approach to feature representation.
While semi-supervised methods exhibit promising performance in the image domain, a notable gap exists in the SSL literature for point clouds, particularly in the context of classification tasks.

\subsection{Semi-supervised Learning in Point Cloud}
Semi-supervised learning has recently been explored by a few prior works on the 3D point cloud classification. For example, \cite{chen2021consistency} introduces a semi-supervised learning framework, leveraging consistency constraints to ensure uniform predictions on both original and perturbed point clouds. This approach mitigates overfitting the limited labelled data available. Moreover, it incorporates pseudo-label generation to provide high-quality labels for unlabelled point clouds, thereby enhancing the supervision of the discriminative model. On the other hand, \cite{chen2021multimodal} presents a multi-modal semi-supervised learning framework featuring an instance-level consistency constraint and a multi-modal contrastive prototype (M2CP) loss. M2CP works towards minimizing intra-class feature variations by optimizing the distance to class prototypes for each object.
ConFid \cite{confid} is another prominent work that specifically addresses class-imbalanced data scenarios, which is a prevalent challenge in the context of point clouds.
ConFid takes into account class-level confidence through a resampling strategy designed to prevent bias towards high-confident classes. Nevertheless, all existing methods fall short of fully harnessing the potential of unlabelled data. This suggests a potential research gap in the realm of SSL-based 3D classification that maximizes the utilization of all available unlabelled data.

\section{Method}
\subsection{Preliminaries on SSL}
Semi-supervised methods learn from a small amount of labelled data and a relatively large amount of unlabelled data in supervised and unsupervised settings, respectively.
Let, $X_{lb} = {(x_i, y_i)_{i=1}^{n}}$ and $X_{ulb} = {(x_i)_{i=1}^{N}}$ represent the labelled and unlabelled sets with $n \ll N$. Here, $x_i \in \mathcal{R}^{m\times 3}$ represents a 3D point cloud with $m$ points, and $y_i$ is the corresponding class label. Consequently, the semi-supervised learning  can be formulated as:

\begin{equation}
    		\min_{\theta}~[
      {\sum_{(x_i,y_i)\in X_{lb}}\mathcal{L}_{sup}(x_i,y_i,\theta)} 
      +\omega   
      {\sum_{x_i\in X_{ulb}}\mathcal{L}_{unsup}(x_i,\theta)}],
    		\label{equ: semiLoss}
    	\end{equation}
Here, $\theta$ represents the learnable parameters, $\mathcal{L}_{sup}$ and $\mathcal{L}_{unsup}$ are supervised loss (cross-entropy loss, $H(x_i, y_i)$) and unsupervised loss.
$\omega$ is the weight that balances the importance of the unsupervised loss.

Unsupervised learning is the key component of semi-supervised learning, exhibiting variations across different SSL methods.
Consistency regularization is one of the common unsupervised losses that is learned by forcing the predictions of the same sample to be consistent under different perturbations (augmentations) \cite{UnMixMatch,mixmatch,remixmatch}. 
Another popular form of unsupervised learning is entropy minimization \cite{fixmatch,flexmatch}, which learns from the unlabelled data by predicting a pseudo-label for each sample. Our method is built over the entropy minimization concept from FixMatch \cite{fixmatch}. More specifically, FixMatch first generates a class prediction for an unlabelled sample from a weakly augmentation of the sample and considers it as a pseudo-label if the confidence of the prediction is above a pre-defined threshold. Finally, these high-confident pseudo-labels are used as supervision for a strong augmentation of the sample. 

Let, $x^{(i)}_{w}=\phi(x_{ulb}^{i})$ and $x_s=\omega(x_{ulb}^{i})$ be the weakly augmented and strongly augmented samples, and $p_m(p_z(x))$ be the output of the model, where, $p_z$ is the encoder and $p_m$ is the classifier. The unsupervised loss of FixMatch can be defined as follows:

\begin{align}
 \mathcal{L}_{unsup}= \frac{1}{\mu B}\sum_{b=1}^{\mu B} 
 \mathbbm{1}
 \big(
 max(p_{m}(p_z(x^{(i)}_{w})))\geq \tau
 \big) \cdot \nonumber \\
H \big( argmax(p_{m}(p_z(x^{(i)}_{w}))), ~p_{m}(p_z(x^{(i)}_{s}))
 \big),
 \label{eq:loss_unsup}
\end{align} 
where $\tau$ is the fixed confidence threshold, $\mu$ is the ratio of unlabelled to labelled batch size, $B$ is the batch size and $H$ represents the cross-entropy loss function. 

Nevertheless, a fixed (high) threshold results in under-utilization of the unlabelled data as the model does not utilize the low-confident samples. To address this, FlexMatch \cite{flexmatch} and ConFid \cite{confid} propose dynamic pseudo labelling to adjust the threshold value considering the class-wise learning status by curriculum learning. In this context, the learning status refers to the ratio of predicted labels to the target label of a class. Specifically, the threshold value is low for high learning status classes to encourage the learning process, and vice-versa. Yet, the model does not utilize all the available unlabelled samples. Following, we describe our proposed solution, AllMatch, which effectively utilizes all the unlabelled samples for learning. 

\subsection{AllMatch}
AllMatch consists of three components, all aimed towards improving the utilization of the unlabelled data: adaptive hard augmentation, inverse learning and contrastive learning. Following, we discuss each of the components in detail.

\vspace{-10pt}
\subsubsection{Adaptive hard augmentation (AHA).}
One limitation of the high confidence-based thresholding method is that the selected (highly confident) samples are already well learned by the model \cite{saa}. As a result, the loss for these samples is very low and, therefore, does not contribute towards the learning of the model. 
Following \cite{saa}, we refer to these high-confident samples as \textit{easy samples} and propose an adaptive hard augmentation (AHA) module to further utilize the easy samples. The basic idea is to apply relatively hard augmentations on the easy samples to introduce further regularization while applying the usual strong augmentations for the remaining samples.
 Note that the alternation between extra hard augmentation and strong augmentation plays a crucial role, as excess augmentation can potentially have a detrimental effect on the model. While the role of hard augmentation is extensively studied in the image semi-supervised learning literature \cite{saa}, it's not well explored in 3D point clouds. 

We define the easy samples based on a historical loss computed until the current epoch $t$. More specifically, we calculate the historical loss for the $i-th$ sample as the exponential moving average (EMA):
\begin{equation}
    \mathcal{H}_{i}^t = (1-\kappa)\cdot\mathcal{H}_i^{t-1}+\kappa \cdot l_i^t,
    \label{eq:historical}
\end{equation}
where, $l_i^t$ is computed as:

\begin{equation}
l_i^t = H \big( argmax(p_{m}(p_z(x^{(i)}_{w}))), ~p_{m}(p_z(x^{(i)}_{s}))
 \big),
 \label{eq:current_loss}
\end{equation}

The lower historical loss for sample $i$ indicates an easy sample on which we apply additional hard augmentations. To divide the samples into easy and hard samples based on the historic loss, we utilize the OTSU \cite{otsu}  function that returns a binary indicator $\mathcal{T}_i$ for each sample. Here, $\mathcal{T}_i=0$ indicates easy sample, and $\mathcal{T}_i=1$ indicates hard sample. Finally, for an easy sample, we apply a series of transformations to act as hard augmentation, including random scaling, rotation, translation and jitters. The AHA module can be represented as:

\begin{equation}
Augmented~sample(x_i)=\left\{\begin{array}{l}
     \mathcal{K}(x_i),\mathcal{T}_i=1\\
     \mathcal{K'}(x_i),\mathcal{T}_i=0
\end{array},
\right.
\label{eq:AHA}
\end{equation}

Here, $\mathcal{K'}(x_i)$ represents the extra hard augmentation, while $\mathcal{K}(x_i)$ represents hard augmentation. In the implementation, we start applying the AHA module after certain epochs of training, defined by `warm-up epochs.' Since the loss is relatively high for all samples at the start of training, adding the AHA from the start of training does not add any benefit. 

\subsubsection{Inverse learning.}
Another challenge in pseudo-labeling-based semi-supervised learning comes from the ambiguity in the predictions of the models. More specifically, the samples with low confidence (high ambiguity) are discarded by the pseudo-labeling concept. These low-confident samples do not contribute to the model's learning process, as pseudo-labels cannot be assigned to them. Consequently, a significant number of unlabelled samples go unused. However, these neglected unlabelled samples can be re-purposed to optimize the model inversely by learning \textit{what not to learn}.

As demonstrated in \cite{fullmatch} for SSL in the image domain, after a certain number of training iterations, the correct classes are situated in one of the top-k predictions (i.e., the $k$ classes with the highest predicted probabilities) of the model. In essence, the model is extremely confident in its predictions for containing the right class in one of these top-k classes. Therefore, it is advantageous to treat the remaining classes (those outside the top-k predictions) as \textit{wrong} class for the input. 
Motivated by \cite{fullmatch}, we incorporate inverse learning, which learns to avoid predicting such classes. 

For Inverse learning, following the convention of FixMatch, we first predict a pseudo-label from a weakly augmented sample and find the value of $k$ for which the model achieves 100\% accuracy on the strongly augmented samples at the current epoch. 
Then, we sort the classes in order of highest to lowest confidence using a ranking function, $Rank$.
Finally, for the classes with $Rank >k$ we perform inverse learning on the strongly augmented samples as:

\begin{equation}
\mathcal{L}_{inv}= - \frac{1}{B} \sum_{i=1}^{B} \sum_{c=1}^{C} \mathbbm{1}[Rank\big(p_{m}(p_z(x^{(i)}_{w}))\big)>k]\ \cdot  \log\big(1-p_{m}(p_z(x^{(i)}_{s}))\big), \label{eq:loss_inv}
\end{equation}

\subsubsection{Contrastive learning (CL).}
While the AHA and inverse learning module increase the utilization of the unlabelled data, the samples that are neither very high-confident nor very low-confident are not utilized by any of the above modules. To ensure learning from the remainder of the sample, we incorporate a contrastive loss on the unlabelled set that learns from all the samples in an unsupervised setting. Contrastive methods learn from positive and negative samples, with the aim of pushing the positive pairs closer in the embedding space and moving them away from the negative sample. In this context, positive samples are the perturbation of the same sample, while all other samples are considered negative. Contrastive learning has shown remarkable success in learning representation in different computer vision domains \cite{simclr,UnMixMatch, jiang2021guided}. Since contrastive learning does not require labels, it ensures total usage of unlabelled data. 

For each unlabelled sample, $x_{ulb}^{(i)} \in X_{ulb}$, a weak perturbation $\phi$ and a strong perturbation $\omega$ is applied to generate a positive pair. In our case, we apply rotation and random scale as weak augmentation, and for strong augmentation, a couple of different augmentations (e.g. jitter, translation, rotation, random scale) are randomly applied. The augmented samples are then fed to the encoder and projection head that generates the embedding $z_i=p_z(y|x_{ulb_i})$, where $[z_{w_i}, z_{s_i}] \in z_i $. Finally, the unsupervised contrastive loss \cite{simclr} function can be described as:

\begin{equation}
\label{eq:loss_con}
    \mathcal{L}_{\text{\textit{con}}} = - \frac{1}{2B} \sum_{i=1}^{2B} \log\frac{\exp(z_i, z_{\kappa(i)}/\tau)}{\sum_{k=1}^{2B} \mathbbm{1}_{[k \neq i]} \exp(z_i, z_k/\tau)} ,
\end{equation} 
where, $\kappa(i)$ is the index of the second augmented sample, $\mathbbm{1}_{[k \neq i]}$ is an indicator function which returns 1 when $k$ is not equal to $i$, and 0 otherwise. $\tau$ is a temperature parameter. Note that the pair samples are the same as the ones used for the unsupervised loss mentioned above. So, the contrastive loss does not increase the computational complexity. 

Additionally, we utilize a modified supervised version of contrastive loss (supervised contrastive loss \cite{supcon}) on the labelled set. Like the unsupervised contrastive loss, the supervised contrastive loss learns from the augmented positive and negative samples. However, the additional label information benefits the model by pushing together samples that belong to the same class. The supervised contrastive loss can be formulated as follows:

\begin{equation}
\label{eq:loss_supcon}
    \mathcal{L}_{\text{\textit{supcon}}} = \sum_{i=1}^{2B} - \frac{1}{2 B_{y_i}-1} \sum_{j=1}^{2B} \mathbbm{1}_{[i \neq j]} \cdot \mathbbm{1}_{[y_i = y_j]} \cdot
    \log\frac{\exp(z_i, z_j/\tau)}{\sum_{k=1}^{2B} \mathbbm{1}_{[k \neq i]} \exp(z_i, z_k/\tau)} ,
\end{equation} 
where, $y_i$ represents class label of sample $i$. $\mathbbm{1}_{[y_i = y_j]}$ is an indicator function that returns 1 when the samples $i$ and $j$ belongs to same class label. ${2 B_{y_i}-1}$ represents number of positive samples from class $y_i$.

\subsection{Total Loss}
Finally, the total loss of AllMatch is the sum of all the described above:
 \begin{equation}
\label{eq:loss_total}
    \mathcal{L} = \mathcal{L}_{\text{\textit{sup}}}+
    \mathcal{L}_{\text{\textit{unsup}}}+\alpha \cdot \mathcal{L}_{\text{\textit{supcon}}}+\beta \cdot \mathcal{L}_{\text{\textit{con}}}+ \gamma \cdot \mathcal{L}_{\text{\textit{inv}}} ,
\end{equation} 
 Here, $\alpha, \beta$ and $\gamma$ are loss factors balancing the importance of each loss. We find optimal values of each factor from the empirical study described in Section \ref{sensitivity}.

 \begin{algorithm*}[t] 
	\caption{AllMatch Algorithm}
	\LinesNumbered 
	\KwIn{Labelled batch $\mathcal{B}_l=\{(x_i,y_i)\}^M$, unlabelled batch $\mathcal{B}_u=\{u_i\}^N$
 }
    Apply AHA to $\mathcal{B}_l$ as Eq. \ref{eq:AHA} \\
    $l_{sup}$ = $\mathcal{L}_{\text{\textit{sup}}}(\mathcal{B}_l)$\\
    $l_{unsup}$ = $\mathcal{L}_{\text{\textit{unsup}}}(\mathcal{B}_u)$ \tcp{via Eq. \ref{eq:loss_unsup}} 
    $l_{con}$ = $\mathcal{L}_{\text{\textit{con}}}(\mathcal{B}_u)$ \tcp{via Eq. \ref{eq:loss_con}} 
    $l_{supcon}$ = $\mathcal{L}_{\text{\textit{supcon}}}(\mathcal{B}_l)$ \tcp{via Eq. \ref{eq:loss_supcon}} 
	\For{$i$ $\leftarrow 1$ to $N$}{
		Compute the rank using, $Rank\big(p_{m}(p_z(x^{(i)}_{w})\big)$
		\For{$c$ $\leftarrow 1$ to $C$}{
            \If {$Rank\big(p_{m}(p_z(x^{(i)}_{w}))\big)> k$} {$l_{inv} += \log\big(1-p_{m}(p_z(x^{(i)}_{s}))\big)$ }
		}
          Compute and update historical loss, $\mathcal{H}_i$ \tcp{via Eq. \ref{eq:historical}}
        Update mark $\mathcal{\tau}_i=\mathbbm{1}(\mathcal{H}_i\le \texttt{OTSU}(H_i))$ 
  }
        $l_{total}$ =  $l_{sup} + l_{unsup} + \alpha \cdot l_{supcon} + \beta \cdot l_{con} + \gamma \cdot l_{inv}$
\end{algorithm*}

\section{Experiments}
In this section, we discuss the experimental results of our study on the ModelNet40 and ScanobjectNN datasets. First, we discuss the main result, followed by an in-depth ablation study of our proposed modules and sensitivity analysis of different module-specific parameters. 

\subsection{Implementation Details}
We evaluate AllMatch on two popular point cloud datasets, ModelNet40 \cite{modelnet} and ScanObjectNN \cite{scanobject}. 
For the implementation details and hyper-parameters, we nearly follow the existing SOTA, ConFid \cite{confid}, for a fair comparison. 
We use PointTransformer \cite{pct} as the encoder. We train the model with SGD optimizer with a learning rate of 0.00005 for 350 epochs. The unlabelled to labelled batch size ratio is 4, and the labelled batch size is 24. 
We utilize rotation and random scale as weak augmentation and random scaling, rotation, translation and jitter as strong augmentation. 
Note that, as a geometric entity, the underline representation of the point clouds must remain invariant under specific transformations \cite{pointnet}. Thus, we only chose the augmentation methods that do not violate this special property of the point clouds. For the AHA module, various combinations of strong augmentation are applied as an additional hard augmentation to the easy samples. A detailed sensitivity analysis on the augmentation strength is discussed in Section \ref{sensitivity}.

\subsection{Main Result}
The overall comparison to the previous works is presented in Tables \ref{tab:Classification-modelnet} and \ref{tab:Classification-scanobject} for ModelNet-40 and ScanObjectNN datasets, respectively. Here, PL \cite{pl}, FixMatch \cite{fixmatch}, Dash \cite{dash}, and FlexMatch \cite{flexmatch} were initially proposed for the image domain and later adopted to the PointCloud, whereas ConFid \cite{confid} is specifically designed for the point cloud. The results for other prior works are directly adopted from the reported results in ConFid \cite{confid}. Following the experimental setups of previous literature on this task, we report the results for different amounts of labelled samples. More specifically, we evaluate the performance on three different sizes of labelled sets: 2\%, 5\%, and 10\% for the ModelNet40 dataset and 1\%, 2\%, and 5\% for the ScanObjectNN dataset. We report the overall accuracy and the mean of class-wise accuracy (mean accuracy) on the test set. 

As we observe from these tables, AllMatch outperforms SOTA by a significant margin across all sizes of unlabelled sets, demonstrating substantial improvement. For example, with only 2\% labelled data from ModelNet40 (Table \ref{tab:Classification-modelnet}), AllMatch outperforms ConFid by a notable 10.48\% and 18.52\% in overall and mean accuracy. While we also see considerable improvements for 5\% and 10\% data, the biggest improvement is observed when the least amount of labelled data (2\%) is utilized. This indicates the effectiveness of AllMatch in learning from a very small amount of labelled data.

\begin{table*}[t]
 \setlength    \tabcolsep{5pt}
\centering
\caption{Comparison of AllMatch with the SOTA methods in 3D object classification task on ModelNet40 dataset with different amounts of labelled sets. Results of the prior works are adopted from \cite{confid}. Here, `Overall Acc' and `Mean Acc' are the overall accuracy and the average of per-class accuracy.}
\vspace{5pt}
 \begin{tabular}{ cc | cc | cc|cc  }
	\hline
        \multicolumn{2}{c|}{\multirow{3}{*}{Method}} &\multicolumn{2}{c|}{2\% } &\multicolumn{2}{c|}{5\% } &\multicolumn{2}{c}{10\% } 
        \\\cline{3-8}
        
         \multicolumn{2}{c|}{} & Overall & Mean& Overall& Mean& Overall& Mean \\

        \multicolumn{2}{c|}{} & Acc & Acc& Acc& Acc& Acc& Acc\\
        \hline
        \multicolumn{2}{c|}{PCT \cite{pct}} & 71.1 & 61.0 & 77.1& 69.2& 84.6& 77.2
        \\ 
        \multicolumn{2}{c|}{PL \cite{pl}} & 69.7 & 59.6 & 78.3& 69.0& 85.1& 77.7
        \\ 
        \multicolumn{2}{c|}{Flex-PL \cite{flexmatch}} & 66.7 & 54.9 & 74.2& 62.3& 83.2& 70.3
        \\ 
        \multicolumn{2}{c|}{ConFid-PL \cite{confid}} & 74.4 & 61.9 & 80.6& 73.5& 86.5& 80.4
        \\ 
        \multicolumn{2}{c|}{FixMatch \cite{fixmatch}} & 70.8 & 62.7 & 78.9& 71.1& 85.5& 79.4
        \\ 
        \multicolumn{2}{c|}{Dash \cite{dash}} & 71.5 & 63.0 & 79.7& 71.8& 85.9& 80.1
        \\
         \multicolumn{2}{c|}{FlexMatch \cite{flexmatch}} & 70.1 & 61.2 & 80.5& 70.4& 86.2& 78.7
        \\
         \multicolumn{2}{c|}{ConFid-Match \cite{confid}} & 73.8 & 64.1 & 82.1& 74.3& 87.8 & 82.5
        \\
        \multicolumn{2}{c|}{\textbf{AllMatch}} & \textbf{84.3} & \textbf{82.6} & \textbf{89.0}& \textbf{87.9}& \textbf{90.2}& \textbf{89.8}
        \\
        \hline
    \end{tabular}

\label{tab:Classification-modelnet}
\end{table*}
Table \ref{tab:Classification-scanobject} illustrates a similar on the ScanObjectNN dataset. With only 1\% of the labelled set, AllMatch outperforms the SOTA by a superior 11.20\% on overall accuracy and 14.67\% on mean accuracy. The maximum performance gain is again observed for the least amount of labelled samples.

\begin{table*}[t!]

\centering
\vspace{5pt}
 \setlength    \tabcolsep{5pt}
 
\caption{Comparison of AllMatch with the SOTA methods in 3D object classification task on ScanObjectNN dataset with different amounts of labelled sets. Results of the prior works are adopted from \cite{confid}. Here, `Overall Acc' and `Mean Acc' are the overall accuracy and the average of per-class accuracy.}

    \begin{tabular}{ cc | cc | cc|cc  }
    
	    \hline
        
        \multicolumn{2}{c|}{\multirow{3}{*}{Method}} 
        &\multicolumn{2}{c|}{1\% } &\multicolumn{2}{c|}{2\% } &\multicolumn{2}{c}{5\% } 
        
        \\\cline{3-8}
         \multicolumn{2}{c|}{} & Overall & Mean& Overall& Mean& Overall& Mean
        \\
        \multicolumn{2}{c|}{} & Acc & Acc& Acc& Acc& Acc& Acc\\
        \hline
        \multicolumn{2}{c|}{PCT \cite{pct}} & 32.1 & 26.1 & 44.7& 36.5& 56.6& 50.0
        \\
        \multicolumn{2}{c|}{PL \cite{pl}} & 31.2 & 25.8 & 47.5& 38.6& 58.1& 51.5
        \\
        \multicolumn{2}{c|}{Flex-PL \cite{flexmatch}} & 29.2 & 24.2 & 47.2& 37.6& 60.1& 51.9
        \\
        \multicolumn{2}{c|}{Confid-PL \cite{confid}} & 32.6& 27.1 & 48.8& 41.5& 63.1& 55.2
        \\
        \multicolumn{2}{c|}{FixMatch \cite{fixmatch}} & 33.5 & 27.6 & 47.4& 39.9& 59.4& 52.4
        \\
         \multicolumn{2}{c|}{Dash \cite{dash}} & 35.1 & 29.3 & 50.3& 44.1& 62.8& 60.3
        \\
        \multicolumn{2}{c|}{FlexMatch \cite{flexmatch}} & 34.2 & 26.2 & 48.5& 39.7& 63.4& 57.2
        \\
         \multicolumn{2}{c|}{ConFid-Match \cite{confid}} & 38.2 & 32.7 & 57.0& 48.6& 69.4& 65.5
        \\
        \multicolumn{2}{c|}{\textbf{AllMatch}} & \textbf{49.4} & \textbf{47.4} & \textbf{60.8}& \textbf{58.9}& \textbf{77.5}& \textbf{76.0}
        
        \\
        \hline
    \end{tabular}

\label{tab:Classification-scanobject}
\end{table*}

\newpage
\subsection{Ablation Study}
\begin{wraptable}{r}{4.5cm}
    \vspace{-35pt}
    \scriptsize
    \setlength    \tabcolsep{5pt}
    \caption{Ablation study of different proposed components of AllMatch on ModelNet40.}
    \vspace{-10pt}
    \begin{center}
    \begin{tabular}{c|c|c|ccc }
    \hline
    \textbf{Inv. lear.} & \textbf{AHA}  & \textbf{CL} & \textbf{OA}\\ \hline
    $\checkmark$ & $\checkmark$ & $\checkmark$  &  \textbf{84.28}  \\
    $\checkmark$ & $\checkmark$ &  & 82.94  \\
    $\checkmark$ &  & $\checkmark$  & 79.98  \\
     & $\checkmark$ & $\checkmark$ & 80.50  \\
     &   & $\checkmark$ &  77.90 \\
    $\checkmark$&   &  & 79.37  \\
     & $\checkmark$ &  & 80.10  \\
      &   &  & 74.80  \\ 
    \hline
    \end{tabular}
    \label{tab:ablation_components}
    \end{center}
    \vspace{-45pt}
\end{wraptable}
In this section, we perform a detailed ablation study to understand the importance of different components in our proposed method. We start with the main ablation on the 3 main components of AllMatch, then perform another ablation on the contrastive learning approach. 

\subsubsection{Main ablation.} 
Table \ref{tab:ablation_components} presents the results for the main ablation study.
Here, we ablate all the combinations of the three main components of AllMatch: adaptive hard augmentation (AHA), inverse learning and contrastive learning (CL). First, we drop these modules individually and then in groups to investigate their significance on overall performance. 
We perform this study with 2\% of the labelled data and report the overall accuracy (OA). 

As we observe from this table, when we drop the AHA module individually, the method's performance declines by 4.29\%. This is considerably higher than the drops observed by dropping inverse learning or contrastive loss, making AHA the most important module in AllMatch. We find the inverse learning to be the second most important component, removing which results in a 3.77\% drop in accuracy. Finally, we find that contrastive learning also plays a critical role. Removing the contrastive learning also results in a 1.34\% drop in accuracy. We find similar trends when two modules are removed at a time. We observe a performance drop of 6.37\% when we remove both the AHA and inverse learning modules. Similarly, removing the AHA and contrastive learning module results in a 4.9\% decline in AllMatch's performance. The least accuracy we find by utilizing a single component is 77.9\% with contrastive learning. Finally, we find that removing all components drastically drops the performance by 74.8\%. This further shows the importance of each of the components in AllMatch.   

\begin{wraptable}{r}{4cm}
    \vspace{-25pt}
    \scriptsize
    \setlength    \tabcolsep{4pt}
    \caption{Ablation study of contrastive loss components of AllMatch on ModelNet40 dataset.}
    \begin{center}
    
    \begin{tabular}{c|c|ccc }
    \hline
    \textbf{Sup.} & \textbf{Unsup.}  & \textbf{OA} \\ \cline{1-3}
    $\checkmark$ & $\checkmark$ & \textbf{84.28}  \\
     & $\checkmark$ & 83.47  \\
    $\checkmark$ &   & 83.06  \\
      &   & 82.94  \\ \hline
    \end{tabular}
    \label{tab:ablation_con}
    \end{center}
    \vspace{-25pt}
    
\end{wraptable}

\subsubsection{Ablation on contrastive learning.}
Next, we perform another study on our contrastive learning module, which consists of supervised and self-supervised contrastive loss. Here, we perform another ablation study by removing the contrastive learning components in different combinations. Similar to the previous study, we do this study on 2\% of labelled data from ModelNet40. The results of this study are presented in Table \ref{tab:ablation_con}. 
As we find from this table, removing both of the components results in a considerable drop in performance, with unsupervised contrastive loss having a greater impact on the final performance. More specifically, when both the supervised and unsupervised contrastive losses are employed, AllMatch achieves its highest performance, 84.28\%. By dropping the unsupervised contrastive loss, the method's overall accuracy drops sharply by 1.21\%. However, when the supervised loss is removed, the overall accuracy declines by 0.8\%. This proves AllMatch is gaining better learning signals from the unsupervised contrastive loss component. Finally, when both contrastive losses are removed, AllMatch returns with AHA and inverse learning setting with an overall loss of 82.94\%.

\subsection{Performance on Increased Labelled Set Size} 
\begin{wrapfigure}{r!}{0.3\textwidth}
\vspace{-25pt}
\begin{center}
    \centering
    \includegraphics[width=0.9\linewidth]{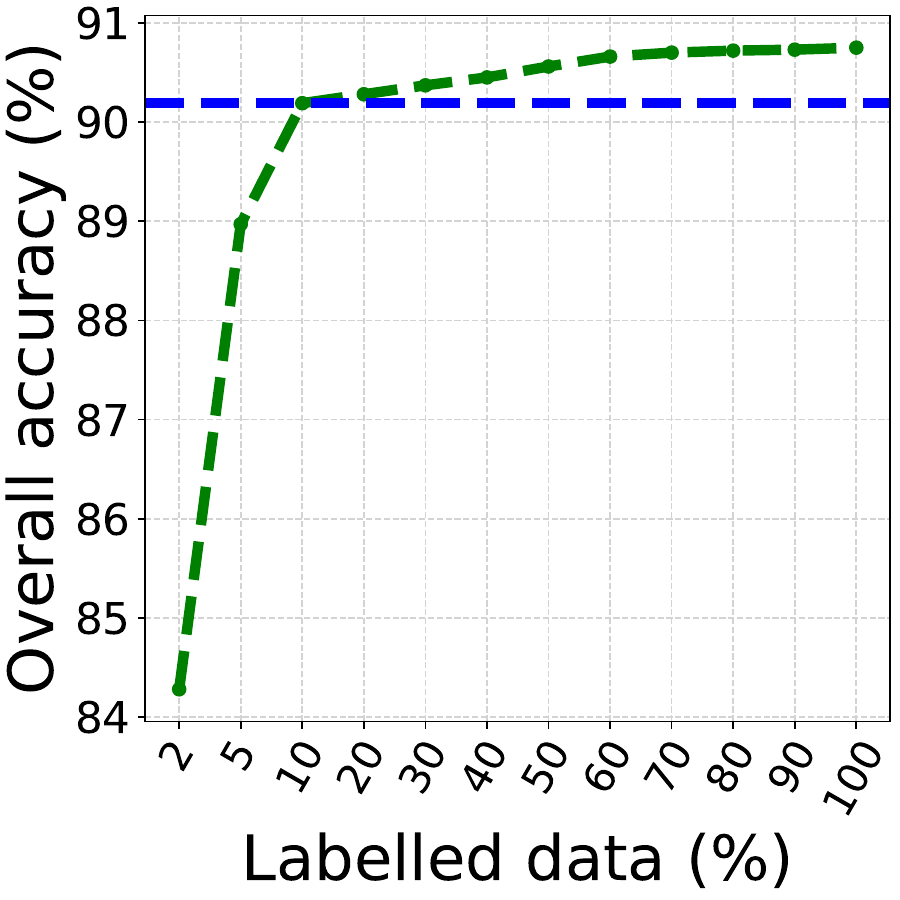}
    \caption{Impact of labelled set size.}
    \label{fig:lb_size}
\end{center}
\vspace{-25pt}
\end{wrapfigure}

In this section, we study the performance of different amounts of labelled data. Specifically, we investigate how an increase in the amount of labelled data impacts performance. To this end, we increase the labelled set size to all labelled data with a 10\% interval (Figure \ref{fig:lb_size}).
It is evident that when only a small amount of labelled data is available (2\% - 5\%), the overall accuracy of AllMatch is relatively low and gradually increases until it reaches 10\% of the labelled set. With 10\% labelled data, the overall accuracy reaches 90.19\%, compared to 90.75\% with the whole labelled set. In other words, the performance with 10\% of labelled samples is almost close to the performance with 100\% of the labelled data. This again proves the fact that AllMatch properly utilizes all the unlabelled data, making it possible for 10\% labelled data to reach such performance.

\subsection{Sensitivity Study} \label{sensitivity}
In this section, we perform detailed sensitivity studies on different parameters of the AllMatch. Following, We discuss the details of these experiments.

\begin{table*}[t]
\caption{Impact of different values of loss factors. }
\resizebox{0.99\linewidth}{!}{
    \centering
    \subfloat[ $\alpha$ vs acc.
        \label{tab:sup_lambda}
    ]{
        \centering
        \begin{minipage}{0.28\linewidth}{
            \begin{center}
            \small
            \begin{tabular}{l c}
                \toprule
                $\alpha$ & Overall accuracy \\
                \midrule
                0.2 &  \textbf{84.28} \\  
                0.5 &  83.38 \\
                1.0 & 81.96 \\ 
                \bottomrule
            \end{tabular}
            \end{center}}
        \end{minipage}
    }
    \hspace{2em}
    \subfloat[$\beta$ vs acc.
        \label{tab:unsup_lambda}
    ]{
    \begin{minipage}{0.28\linewidth}{
        \begin{center}
        \small
        \begin{tabular}{l c}
            \toprule
            $\beta$ & Overall accuracy \\
            \midrule
                0.2 &  \textbf{84.28} \\  
                0.5 &  83.46 \\
                1.0 & 81.24 \\ 
            \bottomrule
        \end{tabular}
        \end{center}}
    \end{minipage}
    }
    \centering
    \hspace{2em}
    \subfloat[$\gamma$ vs acc.
        \label{tab:fullmatch_lambda}
    ]{
    \begin{minipage}{0.28\linewidth}{
        \begin{center}
            \small
            \begin{tabular}{l c}
            \toprule
                $\gamma$ & Overall accuracy \\
            \midrule
                0.2 &  83.22 \\  
                0.5 &  83.38 \\
                1.0 & \textbf{84.28}\\ 
            \bottomrule
            \end{tabular}
        \end{center}}
    \end{minipage}
    }
}
\label{tab:loss_factor}
\vspace{-30pt}
\end{table*}

\subsubsection{Loss factor.} 
The final loss function for training AllMatch consists of multiple losses and includes weight factors to balance the impact of individual loss components. 
Table \ref{tab:loss_factor} illustrates the influence of different loss factors on AllMatch's overall performance. The experiments are conducted on the ModelNet40 dataset with 2\% of labelled data. It is observed that the overall accuracy of AllMatch is sensitive to the values of these factors. Specifically, lower loss factor values lead to higher performances for both supervised and unsupervised contrastive losses, while higher values for the inverse loss factor prove beneficial. This indicates that the inverse loss plays a more significant role in contributing to AllMatch's performance compared to the contrastive loss.

\begin{wrapfigure}{r!}{0.3\textwidth}
\vspace{-25pt}
\begin{center}
    \centering
    \includegraphics[width=0.9\linewidth]{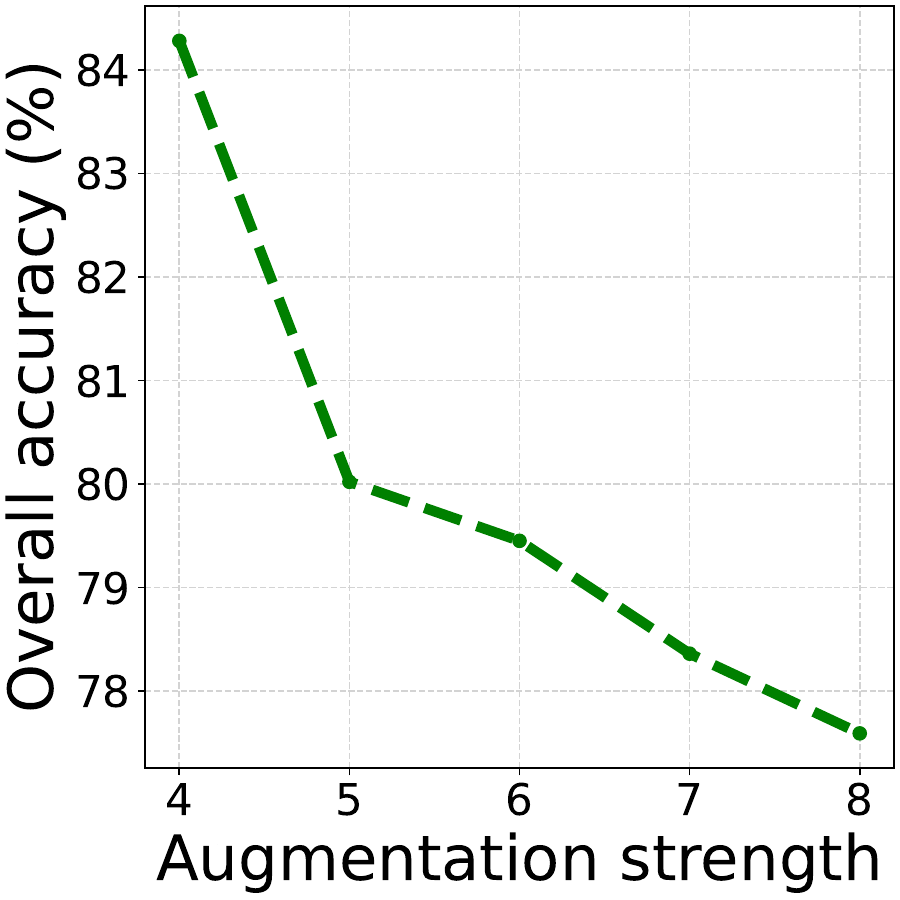}
    \caption{Sensitivity study on the augmentation strengths.}
    \label{fig:aug_strength}
\end{center}
\vspace{-25pt}
\end{wrapfigure}
\subsubsection{Augmentation strength.} 
Augmentation strength is an essential hyperparameter for the Adaptive Hard Augmentation (AHA) module of AllMatch. It indicates the combination of different types of augmentation randomly applied as an additional hard augmentation to the easy samples. In Figure \ref{fig:aug_strength}, we evaluate the overall performance of AllMatch for different augmentation strengths ranging from 4 to 8. Here, the numbers represent the total number of transformations applied to a sample. As we find from Figure \ref{fig:aug_strength}, the best result is observed for using 4 transformations. While the augmentations are essential for learning from easy samples, applying a very high number of augmentations makes the optimization of the model difficult.

\subsubsection{Warm-up epoch for AHA.} 
\begin{wrapfigure}{r!}{0.3\textwidth}
\vspace{-40pt}
\begin{center}
    \centering
    \includegraphics[width=0.9\linewidth]{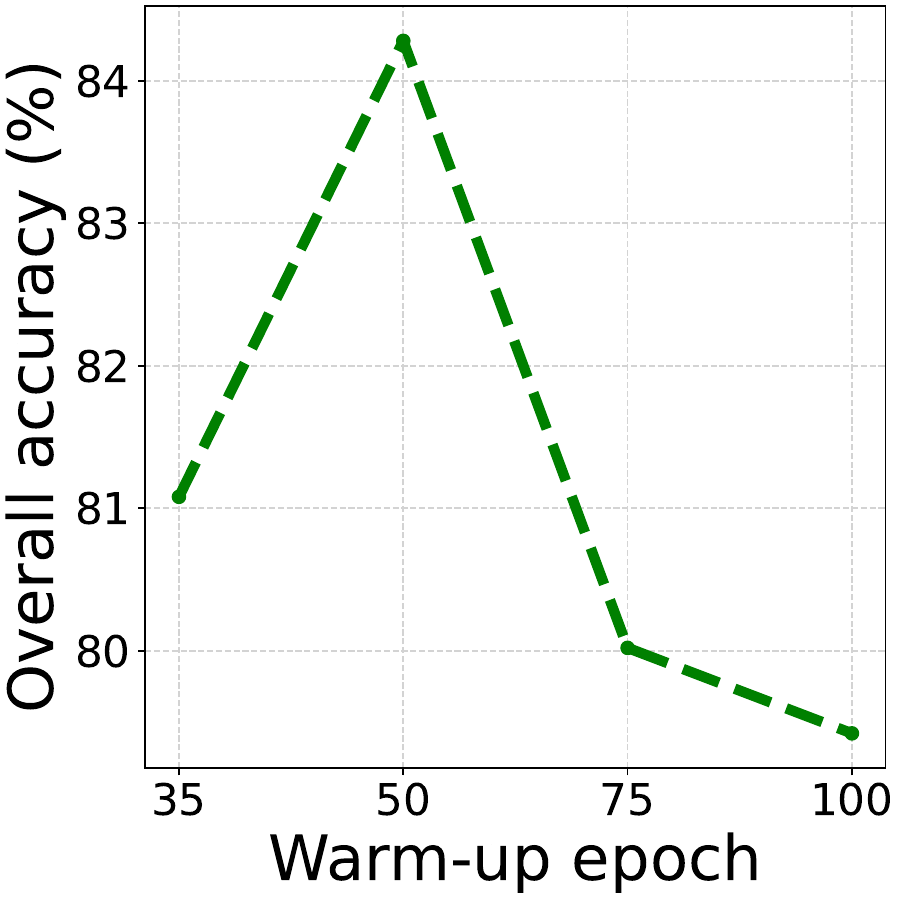}
    \caption{Sensitivity study on the impact of different warm-up epochs.}
    \label{fig:warm_up}
\end{center}
\vspace{-40pt}
\end{wrapfigure}
As discussed in the method section, the AHA module comes into effect after a certain epoch of training. Given that AHA introduces an additional hard augmentation to easy samples, the downstream task becomes challenging for the model. If AHA is applied right at the beginning of the training process, it can negatively impact the model's performance. To this end, warm-up epochs enable the model to learn representations from all samples efficiently during the initial training. As depicted in Figure \ref{fig:warm_up}, AllMatch shows superior performance when warmed up for 50 epochs. However, prolonged warm-up periods diminish performance, as they reduce the effective action period of AHA.

\subsubsection{Epoch and learning rate.} 
\begin{wrapfigure}{r!}{0.5\textwidth}
    \vspace{-25pt}
    \centering
     \begin{subfigure}[b]{0.23\textwidth}
         \centering
         \includegraphics[width=1.\textwidth]{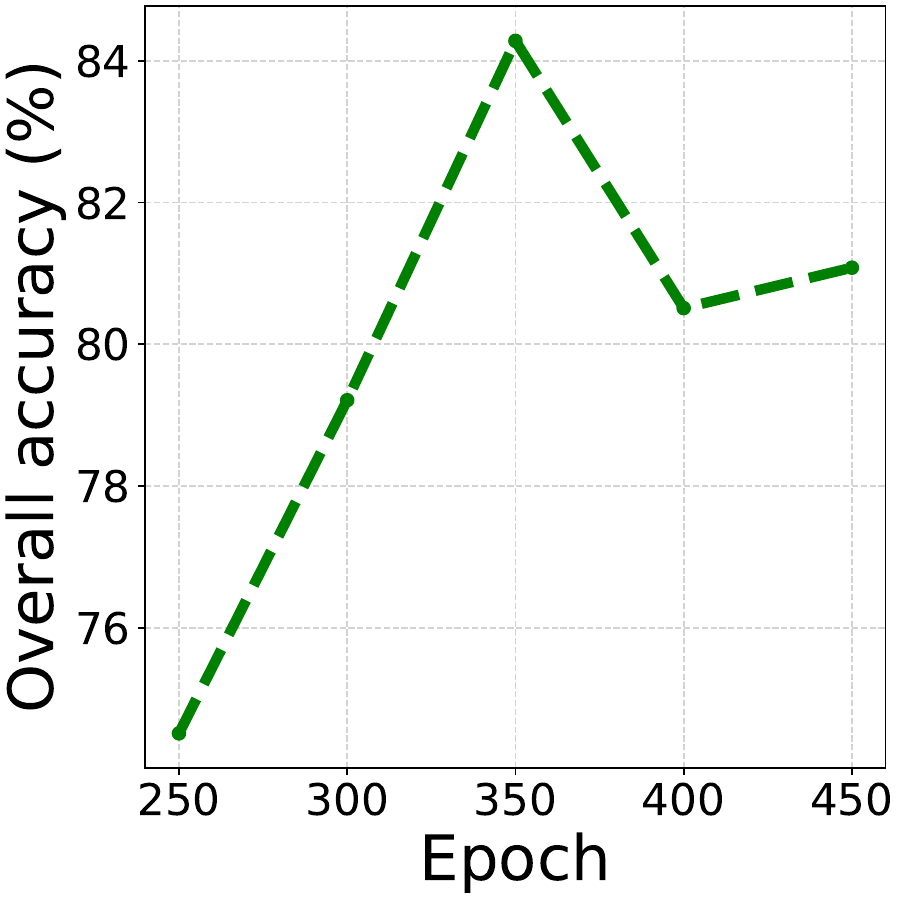}
         \caption{\small ~~Epoch vs Acc.~~}
         \label{subfig:epoch}
     \end{subfigure}
     ~
      \begin{subfigure}[b]{0.23\textwidth}
         \centering
         \includegraphics[width=1\textwidth]{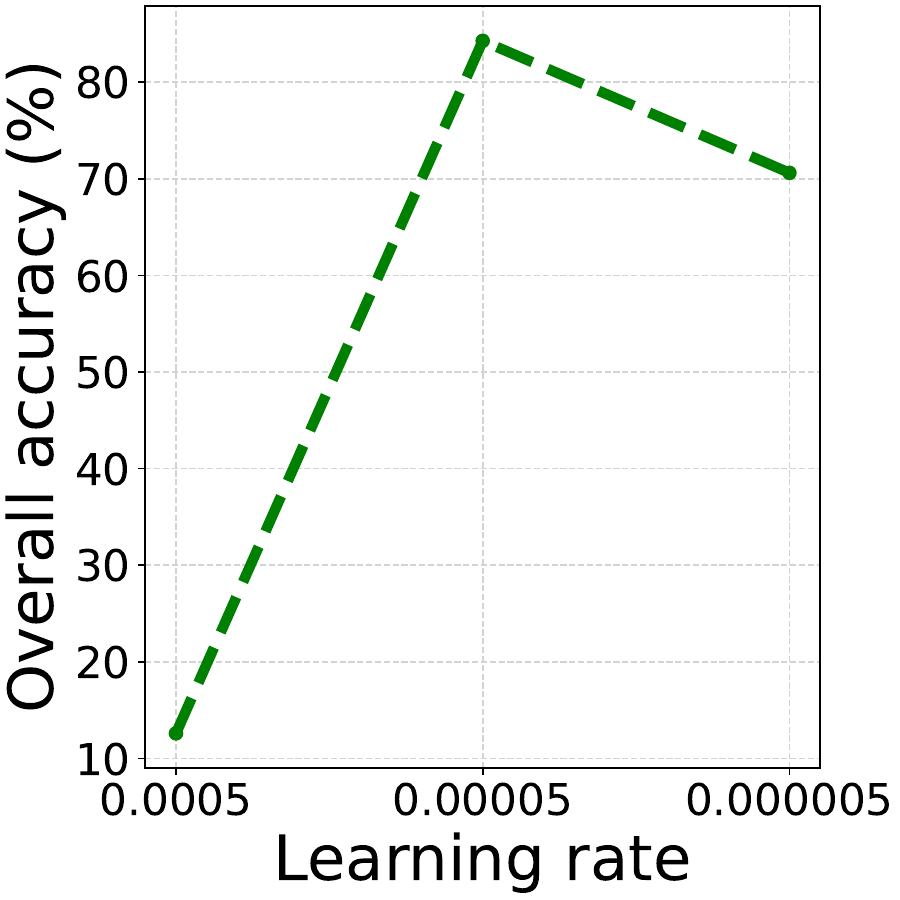}
         \caption{\small LR vs Acc.}
        \label{subfig:lr}
     \end{subfigure}
    \vspace{-1em}
    \caption{Sensitivity study of epoch and learning rate on ModelNet40. 
    }
    \label{fig:analysis_stability}
\vspace{-20pt}
\end{wrapfigure}
Finally, we study two hyper-parameters that are important for the optimization of the model, epochs and learning rate. As depicted in Figure \ref{subfig:epoch}, we observe that AllMatch's performance improves by increasing the number of epochs, reaching its peak at 350 epochs with a best performance of 84.28\%. However, extending the number of epochs beyond this point results in a decline in performance. This is considerably lower than the previous SOTA, which requires 500 epochs of training. This further shows the benefit of utilizing the unlabelled data properly. In Figure \ref{subfig:lr}, we analyze the impact of different learning rates on AllMatch's performance. Notably, lower learning rates contribute to improved performance, whereas higher values lead to a collapse in the model's performance. For instance, at a learning rate of 0.0005, AllMatch achieves only 12\% overall accuracy, emphasizing the sensitivity of the model's performance to the choice of learning rate.

\section{Conclusion}

In this work, we identify that existing 3D semi-supervised techniques fail to capitalize the whole unlabelled data. To solve this issue, we introduce AllMatch, a novel semi-supervised method for point cloud classification that effectively utilizes the entire unlabelled set to improve the learnt representation. Comprised of adaptive hard augmentation, inverse learning, and contrastive learning components, AllMatch outperforms existing SOTA methods by a significant margin, showcasing superior performance even with minimal (1\%) labelled sets. Remarkably, AllMatch, with just 10\% of the labelled set, attains a performance close to the fully supervised learning with all samples, underscoring its efficiency in minimizing reliance on labelled data. 

\noindent\textbf{Limitations.}
While our proposed method has made significant improvement in efficiently leveraging the entire unlabelled set, there are still opportunities for further exploration in this promising direction. In the AHA module, we successfully incorporated selected transformation-invariant augmentation techniques, showcasing a positive step forward. Expanding our exploration to encompass a more diverse range of augmentation methods could prove to be a fruitful avenue for additional advancements. Furthermore, we only explored contrastive learning as the self-supervised module to ensure the utilization of all unlabelled samples, leaving room for exploring different versions and alternatives of contrastive loss.
\clearpage  
\bibliographystyle{splncs04}
\bibliography{egbib}

\begin{thebibliography}{10}
\providecommand{\url}[1]{\texttt{#1}}
\providecommand{\urlprefix}{URL }
\providecommand{\doi}[1]{https://doi.org/#1}

\bibitem{remixmatch}
Berthelot, D., Carlini, N., Cubuk, E.D., Kurakin, A., Sohn, K., Zhang, H., Raffel, C.: Remixmatch: Semi-supervised learning with distribution matching and augmentation anchoring. In: ICLR (2019)

\bibitem{mixmatch}
Berthelot, D., Carlini, N., Goodfellow, I., Papernot, N., Oliver, A., Raffel, C.A.: Mixmatch: A holistic approach to semi-supervised learning. NeurIPS  (2019)

\bibitem{softmatch}
Chen, H., Tao, R., Fan, Y., Wang, Y., Wang, J., Schiele, B., Xie, X., Raj, B., Savvides, M.: Softmatch: Addressing the quantity-quality tradeoff in semi-supervised learning. In: ICLR (2022)

\bibitem{chen2021consistency}
Chen, L., Zhang, Y., Lin, Y., Jiang, M., Huang, Y., Lei, Y.: Consistency-based semi-supervised learning for point cloud classification. In: PRAI. pp. 440--445 (2021)

\bibitem{simclr}
Chen, T., Kornblith, S., Norouzi, M., Hinton, G.: A simple framework for contrastive learning of visual representations. In: ICML. pp. 1597--1607 (2020)

\bibitem{fullmatch}
Chen, Y., Tan, X., Zhao, B., Chen, Z., Song, R., Liang, J., Lu, X.: Boosting semi-supervised learning by exploiting all unlabeled data. In: CVPR. pp. 7548--7557 (2023)

\bibitem{chen2021multimodal}
Chen, Z., Jing, L.: Multimodal semi-supervised learning for 3d objects. In: BMVC (2021)

\bibitem{confid}
Chen, Z., Jing, L., Yang, L., Li, Y., Li, B.: Class-level confidence based 3d semi-supervised learning. In: WACV. pp. 633--642 (2023)

\bibitem{saa}
Gui, G., Zhao, Z., Qi, L., Zhou, L., Wang, L., Shi, Y.: Enhancing sample utilization through sample adaptive augmentation in semi-supervised learning. In: CVPR. pp. 15880--15889 (2023)

\bibitem{pct}
Guo, M.H., Cai, J.X., Liu, Z.N., Mu, T.J., Martin, R.R., Hu, S.M.: Pct: Point cloud transformer. Computational Visual Media  \textbf{7},  187--199 (2021)

\bibitem{jiang2021guided}
Jiang, L., Shi, S., Tian, Z., Lai, X., Liu, S., Fu, C.W., Jia, J.: Guided point contrastive learning for semi-supervised point cloud semantic segmentation. In: ICCV. pp. 6423--6432 (2021)

\bibitem{supcon}
Khosla, P., Teterwak, P., Wang, C., Sarna, A., Tian, Y., Isola, P., Maschinot, A., Liu, C., Krishnan, D.: Supervised contrastive learning. NeurIPS pp. 18661--18673 (2020)

\bibitem{conmatch}
Kim, J., Min, Y., Kim, D., Lee, G., Seo, J., Ryoo, K., Kim, S.: Conmatch: Semi-supervised learning with confidence-guided consistency regularization. In: ECCV. pp. 674--690 (2022)

\bibitem{pl}
Lee, D.H., et~al.: Pseudo-label: The simple and efficient semi-supervised learning method for deep neural networks. In: Workshop on challenges in representation learning, ICML. p.~896 (2013)

\bibitem{comatch}
Li, J., Xiong, C., Hoi, S.C.: Comatch: Semi-supervised learning with contrastive graph regularization. In: ICCV. pp. 9475--9484 (2021)

\bibitem{liu2022point}
Liu, F., Lin, G., Foo, C.S., Joshi, C.K., Lin, J.: Point discriminative learning for data-efficient 3d point cloud analysis. In: IEEE Int. Conf. on 3D Vision. pp. 42--51 (2022)

\bibitem{sequencematch}
Nguyen, K.B.: Sequencematch: Revisiting the design of weak-strong augmentations for semi-supervised learning. In: WACV. pp. 96--106 (2024)

\bibitem{otsu}
Otsu, N.: A threshold selection method from gray-level histograms. TSMC  \textbf{9},  62--66 (1979)

\bibitem{pointnet}
Qi, C.R., Su, H., Mo, K., Guibas, L.J.: Pointnet: Deep learning on point sets for 3d classification and segmentation. In: CVPR. pp. 652--660 (2017)

\bibitem{UnMixMatch}
Roy, S., Etemad, A.: Scaling up semi-supervised learning with unconstrained unlabelled data. In: AAAI. pp. 14847--14856 (2024)

\bibitem{fixmatch}
Sohn, K., Berthelot, D., Carlini, N., Zhang, Z., Zhang, H., Raffel, C.A., Cubuk, E.D., Kurakin, A., Li, C.L.: Fixmatch: Simplifying semi-supervised learning with consistency and confidence. NeurIPS pp. 596--608 (2020)

\bibitem{mean_teacher}
Tarvainen, A., Valpola, H.: Mean teachers are better role models: Weight-averaged consistency targets improve semi-supervised deep learning results. In: NeurIPS. pp. 1195--1204 (2017)

\bibitem{scanobject}
Uy, M.A., Pham, Q.H., Hua, B.S., Nguyen, D.T., Yeung, S.K.: Revisiting point cloud classification: A new benchmark dataset and classification model on real-world data. In: ICCV. pp. 1588--1597 (2019)

\bibitem{chmatch}
Wu, J., Yang, H., Gan, T., Ding, N., Jiang, F., Nie, L.: Chmatch: Contrastive hierarchical matching and robust adaptive threshold boosted semi-supervised learning. In: CVPR. pp. 15762--15772 (2023)

\bibitem{modelnet}
Wu, Z., Song, S., Khosla, A., Yu, F., Zhang, L., Tang, X., Xiao, J.: 3d shapenets: A deep representation for volumetric shapes. In: CVPR. pp. 1912--1920 (2015)

\bibitem{uda}
Xie, Q., Dai, Z., Hovy, E., Luong, T., Le, Q.: Unsupervised data augmentation for consistency training. In: NeurIPS. pp. 6256--6268 (2020)

\bibitem{dash}
Xu, Y., Shang, L., Ye, J., Qian, Q., Li, Y.F., Sun, B., Li, H., Jin, R.: Dash: Semi-supervised learning with dynamic thresholding. In: ICML. pp. 11525--11536 (2021)

\bibitem{flexmatch}
Zhang, B., Wang, Y., Hou, W., Wu, H., Wang, J., Okumura, M., Shinozaki, T.: Flexmatch: Boosting semi-supervised learning with curriculum pseudo labeling. NeurIPS pp. 18408--18419 (2021)

\bibitem{simmatch}
Zheng, M., You, S., Huang, L., Wang, F., Qian, C., Xu, C.: Simmatch: Semi-supervised learning with similarity matching. In: CVPR. pp. 14471--14481 (2022)

\end{thebibliography}
\end{document}